\documentclass[10pt,twocolumn,letterpaper]{article}

\usepackage[pagenumbers]{cvpr}      

\usepackage[dvipsnames]{xcolor}

\definecolor{cvprblue}{rgb}{0.21,0.49,0.74}
\usepackage[pagebackref,breaklinks,colorlinks,citecolor=cvprblue]{hyperref}
 \usepackage{mathrsfs}
 \usepackage{multirow}
\def\paperID{13862} 
\def\confName{CVPR}
\def\confYear{2024}

\title{Steal My Artworks for Fine-tuning? A Watermarking Framework for Detecting Art Theft Mimicry in Text-to-Image Models}

\author{Ge Luo, Junqiang Huang, Manman Zhang, Zhenxing Qian, Sheng Li, Xinpeng Zhang\\
School of Computer Science, Fudan University\\}

\begin{document}
\maketitle

\begin{abstract}

The advancement in text-to-image models has led to astonishing artistic performances. However, several studios and websites illegally fine-tune these models using artists' artworks to mimic their styles for profit, which violates the copyrights of artists and diminishes their motivation to produce original works. Currently, there is a notable lack of research focusing on this issue. In this paper, we propose a novel watermarking framework that detects mimicry in text-to-image models through fine-tuning.
This framework embeds subtle watermarks into digital artworks to protect their copyrights while still preserving the artist's visual expression. If someone takes watermarked artworks as training data to mimic an artist's style, these watermarks can serve as detectable indicators. By analyzing the distribution of these watermarks in a series of generated images, acts of fine-tuning mimicry using stolen victim data will be exposed. In various fine-tune scenarios and against watermark attack methods, our research confirms that analyzing the distribution of watermarks in artificially generated images reliably detects unauthorized mimicry.

\end{abstract}    
\section{Introduction}
\label{sec:intro}

\begin{figure}[htb]
  \includegraphics[width=0.46\textwidth]{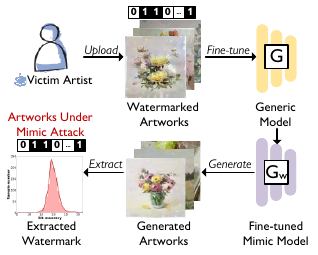}
  \caption{Artists who upload their works online are vulnerable to stylistic imitation. We are employing watermarking techniques to detect and expose unauthorized fine-tuning in suspect models or services, thereby safeguarding the originality of these artworks. }
  \label{fig:first}
\end{figure}

Generative imagery technology has revolutionized the field of art, making it accessible and versatile. Applications such as Stable Diffusion\cite{webui}, Midjourney\cite{mid}, and DALL-E-3\cite{dalle3} enable individuals with no artistic background to create art pieces from simple descriptions. This democratization of art creation extends to experienced artists as well, who can leverage these tools to generate multiple pieces for inspiration or to revise their existing work. The introduction of generative models has simplified the traditionally labor-intensive process of artistic creation, allowing artists to focus on innovation rather than execution. These models have found applications across diverse fields, enriching video game design, book illustrations, magazine covers, and presentations with AI-crafted art. Remarkably, these models can be fine-tuned to emulate the unique styles of specific artists, producing new works that maintain the original's aesthetic qualities. Additionally, online platforms now facilitate the sharing of these fine-tuned models, enabling the replication of artist styles, celebrity likenesses, and even the creation of NSFW content.

The prowess of artificial intelligence in creating compelling artworks is evident, as demonstrated by several pieces that have secured awards in various competitions. However, AI image generation models come with a set of significant issues. These include copyright~\cite{brittain2023ai}, ethics, and consent. A new area of debate is the fine-tuning of text-to-image models to replicate the unique styles of established artists, an action that has led to considerable negative consequences for the individuals affected. In an interview addressing an instance of this kind of imitation, the artist Hollie Mengert, who was directly impacted, shared her insights~\cite{baio2022invasive}: "As far as the characters, \underline{I didn’t see myself in it}. I didn’t personally see the AI making decisions that I would make, so I did feel distance from the results. Some of that frustrated me because it feels like it isn’t actually mimicking my style, and \underline{yet my name is still part of the tool}." 

The issues arising from the replication of painting styles are widespread, affecting a multitude of artists, with Hollie Mengert being just one notable example. Firstly, the proliferation of imitations hinders the efforts of original artists to promote their work and cultivate their personal brands, often leading to their obscurity due to the deluge of copied art, which can significantly undermine their earnings and livelihood. Secondly, this replication poses a discouraging challenge to aspiring artists and hobbyists, whose passion and future dedication to crafting a distinct style may be easily overshadowed by the capabilities of AI to replicate such styles. This could lead to a discouragement that dissuades them from honing their craft. Thirdly, the application of finely-tuned mimicry in contexts such as NSFW content, or illustrations promoting racial discrimination, endangering nations, or disrupting peace, could be deleterious to the original artists' reputation.

In this paper, we focus on protecting artists' rights and aiding regulatory bodies in identifying unauthorized use of artwork in fine-tuning generative models. We utilize non-intrusive watermarking technology that respects original artistic expression and work promotion. Our framework is tailored to address fine-tuning imitations in generative art, allowing artists to embed unique watermarks into their artwork before online sharing. These imperceptible watermarks preserve visual integrity and, when used in fine-tuning, become part of the training process, revealing any unauthorized use in generated artworks. We extensively test our framework in various fine-tuning scenarios, including solo and collaborative imitations and mixing imitations with clean images, demonstrating its robustness. Additionally, we evaluated our watermarking framework's resilience against traditional watermark attacks and challenges arising from fine-tuning imitations.

Our summarized contributions are:
\begin{itemize}
\item We propose a watermarking framework, empowering artists to embed unique watermarks in their works for detecting plagiarism caused by mimic fine-tuned models.

\item We design a verification mechanism to discern watermark distributions in outputs from stolen and clean models, enhancing the identification of plagiarism.

\item We verify the effectiveness of this framework in diverse mimicry scenarios, including both single and multi-artist mimicry, ensuring reliable theft detection.

\item We conduct comprehensive tests under both traditional and novel attack scenarios,  confirming the effectiveness and robustness of our framework.
\end{itemize}
\section{Related Work}
\label{sec:relat}
\noindent\textbf{Text-to-Image Generation and Diffusion Models}.
The expressive power of visuals often surpasses text, driving the pursuit of generating visual imagery from text descriptions in academic and industrial realms. This field has seen diverse research methods yielding impressive results~\cite{mansimov16_text2image, reed2016generative, zhang2017stackgan,xu2018attngan, li2019controllable, ramesh2021zero,ding2021cogview,wu2022nuwa, yu2022scaling, nichol2022glide, saharia2022photorealistic, rombach2022high, ramesh2022hierarchical}.GAN-based methods\cite{reed2016generative,zhang2017stackgan, xu2018attngan,li2019controllable} showed promise on smaller datasets. The introduction of autoregressive methods, particularly with DALL-E~\cite{ramesh2021zero} and Parti~\cite{yu2022scaling}, transformed the field, especially in large-scale data settings. Recently, a shift towards diffusion models~\cite{nichol2022glide, saharia2022photorealistic,rombach2022high, ramesh2022hierarchical} has occurred, now the mainstream trend. These models are favored for stable output, superior image quality, and adaptable control, becoming the preferred choice for various applications. Tools like Sd-Webui~\cite{webui} have emerged, helping users share models and artistic works online~\cite{civitai}. While text-to-image models have significantly advanced artistic creation, they also pose challenges to artists and the art community.

\begin{figure*}[thb]
  \includegraphics[width=\textwidth]{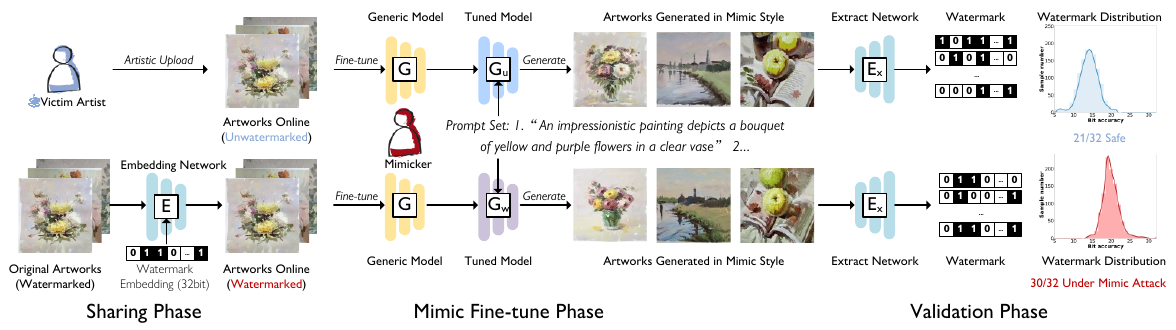}
  \caption{The overview of our watermarking framework that facilitates the embedding of visually non-intrusive watermarks within artworks, which remain extractable following sophisticated imitative fine-tuning. The pattern of watermark retrieval from these generated items enables the detection of potential unauthorized use of artists' works by imitation models. Upon embedding a 32-bit watermark, our system can achieve a bit accuracy rate as high as 93.75\% (30/32).}
  \label{fig:overview}
\end{figure*}

\noindent\textbf{Fine-tune for Art Style Mimicry}. Style mimicry typically involves neural networks producing artworks that mimic a particular artist's style or a distinct artistic genre. Training a model for specific style mimicry requires a considerable collection of artworks in that style. However, with a scarcity of training samples, mastering the nuances of the style and subtle artistic habits becomes challenging. Fine-tuning, a mature technique in fields like natural language processing and computer vision, is also beneficial for style mimicry. Using tools such as Dreambooth~\cite{ruiz2023dreambooth} and Lora~\cite{hu2021lora}, users need only about 20 examples of the target style, along with a pre-trained general model, to effectively mimic the style. Models fine-tuned for style mimicry and their artworks have gained popularity on various art-sharing websites~\cite{civitai,mid}. However, this mimicry often occurs without the original artists' consent, which may infringe upon their rights.

\noindent\textbf{Protecting Against Mimicry Attacks}.
Artists often struggle to prevent unauthorized online distribution of their works, which are frequently used for training and mimicry attacks. Halting this sharing can significantly reduce their promotional efforts. Recent research combats this by introducing invisible perturbations into artworks using adversarial sample technology, impeding fine-tuning training and model replication~\cite{glaze}. Fine-tuning for style mimicry, crucial in advancing computer vision and image generation, requires extensive data, highlighting concerns about artists' rights. Unlike these methods, our approach safeguards artists' rights through watermarking technology, confirming ownership of fine-tuning data in anonymous models and managing rights authorization in signed models.

\section{Treat Model}
\label{sec:pre}

\subsection{Artists}
\noindent\textbf{Artists Goal}. Artists showcase artworks on online platforms for promotion, commissions, and income. They seek to verify if unattributed services or models use their data for imitation, or if signed models have authorization.

\noindent\textbf{Artists Capabilities}.
Our assumption is that the artist possesses the following capabilities: 
\begin{itemize}
\item Artists have the option to insert watermarks signifying personal data or non-authorization into their works before online sharing, maintaining visual integrity.
\item Artists can utilize adequate computational resources or enlist watermarking agencies for watermark application and later validation. 
\item Unaware of mimickers' image caption tools, authors can try various content descriptions as prompts.
\end{itemize}

\subsection{Mimic Attackers}
\noindent\textbf{Attack Goal}.
The mimic attack fine-tunes a text-to-image model to replicate a specific artist's style, the victim. Once fine-tuned, it generates artworks mimicking the victim's style, creating new content of the target style.

\noindent\textbf{Potential attackers}.
Potential initiators of imitation attacks could encompass a range of entities with interests in specific artistic styles or the artists themselves. These may include AI and internet companies, studios specializing in AI, fine arts, and media, as well as web platforms focused on art and painting. Alternatively, these initiators might be individuals particularly interested in the style or the artists.

\noindent\textbf{Attacker Capabilities}.
Our assumption is that the attacker possesses the following capabilities: 
\begin{itemize}
\item Ability to access and download targeted artist's artwork.
\item The capacity to obtain a high-quality pre-trained text-to-image model, such as the SD2.1 checkpoint.
\item Sufficient computational resources and the requisite technical expertise to conduct fine-tuning training.
\end{itemize}

\section{Watermark Framework}
\label{sec:frame}
\begin{figure*}[thb]
  \includegraphics[width=\textwidth]{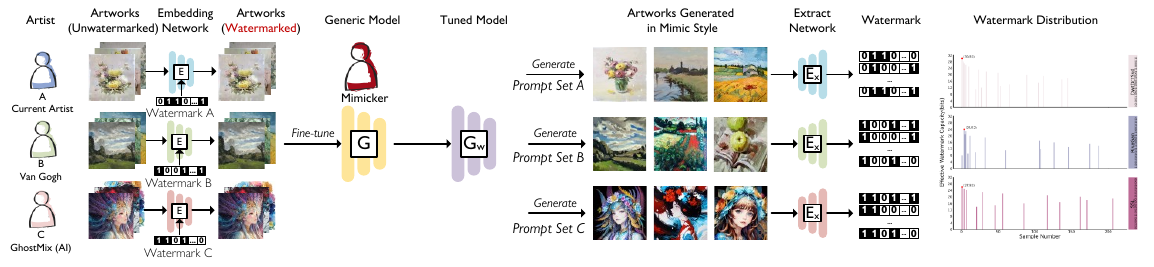}
  \caption{Multi-Artist fine-tuning scenario. They may use the same watermarking algorithm with different watermarks, or employ different watermarking algorithms with varying watermarks (this figure illustrates different method).}
  \label{three}
\end{figure*}

\subsection{Overview}
Figure \ref{fig:overview} illustrates our proposed watermarking framework, which is segmented into three phases: Artist Sharing, mimicry Fine-Tuning, and Verification Extraction. In the Artist Sharing phase, artists integrate watermarks into their works. Imitators then use these watermarked artworks for fine-tuning training, aiming to replicate the artist's style in a generative model. During the Verification Extraction phase, if victims or authenticating bodies identify suspect models, they can interrogate these models, either by accessing the generation service or by downloading the model. This interrogation involves extracting watermarks from various generated artworks, a process that clearly identifies unauthorized fine-tuning theft.

\subsection{General Methodology}
\noindent\textbf{Embedding}.
Watermark embedding is primarily the process of integrating watermark information into images in a manner that is imperceptible to the naked eye, thereby maintaining the image's visual integrity. This approach is equally effective for artworks, where the embedded watermarks, though minutely concealed, do not alter the artwork's artistic expression. The typical steps involved in the watermark embedding process are as follows:
\begin{equation}\label{Gc}
\mathit{I_w}=\mathit{E(\mathit{I, bitstream})}
\end{equation}
where $E$ is the watermark embedding network. Watermark information is encoded as a bitstream. $I$ denotes original images without watermarks, and $I_w$ refers to the watermarked image collection.

\noindent\textbf{Mimic Fine-tune}.
Companies or individuals with an interest in an artist's style can utilize a select set of the artist's works along with a pre-trained generic text-to-image model, $G$, for fine-tuning. This fine-tuning process results in a tuned model, either $G_u$ when fine-tuned with the watermarked images collection $I_w$, or $G_u$ when using the original, unwatermarked images collection $I$, specifically adapted to the target style:
\begin{equation}\label{F1}
G_u=\mathcal{F}(G,I),
\end{equation}
\begin{equation}\label{F2}
G_w=\mathcal{F}(G,I_{w}).
\end{equation}
With $G_u$ and $G_w$, mimickers and users – either through accessing services provided by mimickers or by downloading shared models – have the ability to create artworks $I_s$ in the style of the impacted artist. This creation is facilitated by using content prompts $P_{con}$ and special tags $P_{sp}$, which are defined by imitators during the fine-tuning process. These tags, typically words associated with the artist or their artistic style, are usually released alongside the model:
\begin{equation}\label{Gs}
I_s=G_{w}(\mathit{P_{con}+P_{sp}}).
\end{equation}

\noindent\textbf{Extraction}.
By using a watermark extractor $E_x$, which is specifically aligned with the algorithm, we can successfully extract the watermark $\bar{w} $ from a generated artifact $i_s$:
\begin{equation}\label{Gs}
\bar{w}=E_{x}(\mathit{i_s}).
\end{equation}

\noindent\textbf{Verification}.
A common validation approach is to compare the watermark data w derived from outputs produced by a potentially untrustworthy application service against the author's embedded watermark, distinguishing between unauthorized watermark $w_{ua}$ and authorized watermark $w_{a}$. Bit accuracy is then evaluated to ascertain if the mimicry involves fine-tuning of the author's work and the legitimacy of the used work:
\begin{equation}\label{Gs}
acc=\frac{\bar{n} }{n_{ua}},or,acc=\frac{\bar{n} }{n_{a}}
\end{equation}
where n is the count of correctly extracted watermark bits, with $n_{ua}$
and $n_a$ representing the total bits in unauthorized and authorized watermarks, usually 32. The watermark bit accuracy of a single image lacks persuasive power in indicating whether the victim's work was used for fine-tuning. We identify potential data theft for mimicry fine-tuning by analyzing watermark accuracy extremes, averages, and distributions across outputs from multiple suspect models.

\subsection{Fine-Tuning Scenarios}
\noindent\textbf{Single-Artist}.
The most common form of fine-tuning involves collecting works from a single artist or a singular style for mimicry purposes. This method of fine-tuning is cost-effective and achieves the desired mimicry effects efficiently. This is introduced in the general method section of the previous subsection.

\noindent\textbf{Multi-Artist}.
In figure~\ref{three}, Another scenario involves mimickers adopting a hybrid fine-tuning approach with works from various authors, potentially for stylistic resemblances. In our framework, artists embed unique watermarks in their creations. During verification, content prompts from each artist's work extensively generate respective artworks. Watermarks are then extracted and analyzed for distribution, determining if any artist has suffered from unauthorized fine-tuning mimicry.

\noindent\textbf{Mixed Clean Artworks}.
In a likely scenario, mimickers would gather artworks, most of which are probably without watermarks, except for the ones belonging to the artists being mimicked. This situation is comparable to mixing these with unmarked images for fine-tuning on a moderate scale. To assess this, we conducted tests using different ratios of images with watermarks in the fine-tuning process. The results and detailed verification outcomes of these tests are reported in the experimental section of this paper.

\begin{table*}[]
\resizebox{\linewidth}{!}{
\centering
\begin{tabular}{ccccccccc}
\hline
                        &                             & \multicolumn{5}{c}{{Regarding the sample numbers within an accurate bit range}}                                                               &                         &                             \\ \cline{3-7}
\multirow{-2}{*}{{ 32-bit}} & \multirow{-2}{*}{{watermarked}} & { 0-20\%} & {20-40\%} & { 40-60\%} & { 60-80\%} & { 80-100\%} & \multirow{-2}{*}{{ avg(bits)}} & \multirow{-2}{*}{best(bits)} \\ \hline
                                               & $\times$                                             & 0                             & 255                            & 741                            & 4                              & 0                               & 14.03                                             & 21                          \\
\multirow{-2}{*}{Artists' images}              & $\checkmark$                                         & 0                             & 0                              & 109                            & 867                            & 24                              & 19.54                                             & 29                          \\ \hline
                                               & $\times$                                             & 0                             & 163                            & 826                            & 11                             & 0                               & 14.66                                             & 20                          \\
\multirow{-2}{*}{AIs' images}                  & $\checkmark$                                         & 0                             & 0                              & 193                            & 797                            & 10                              & 19.07                                             & 28                          \\ \hline
                                               & $\times$                                             & 0                             & 176                            & 805                            & 10                             & 0                               & 14.47                                             & 21                          \\
\multirow{-2}{*}{Natural images}               & $\checkmark$                                         & 0                             & 0                              & 520                            & 468                            & 12                              & 19.31                                             & 28                          \\ \cline{1-9} 
\end{tabular}}
\caption{ Average results of watermark (32bit) extraction from 1000 images, generated via single-artist fine-tuning, across 10 groups.\label{30}}
\end{table*}

\subsection{Watermark Attack}
We have comprehensively considered several forms of watermark attacks that may occur in the context of mimicry fine-tuning (assuming the mimicker is knowledgeable about watermark attack techniques), and the validation results are reported in the experimental section.

\noindent\textbf{Traditional Watermark Attack}.
In traditional watermark attacks, mimickers simply process the collected artworks and then use the processed images for fine-tuning mimicry. 

\noindent\textbf{Secondary Fine-Tuning Attack}.
In mimicry scenarios, a mimic may choose two fine-tuning iterations. The first round produces high-quality works, used in subsequent iterative fine-tuning rounds. This continues until the desired effect and consistent output quality are achieved.

\noindent\textbf{Watermark Overlay Attack}. Watermark overlay is a classic issue in watermark attacks. We considered the scenario where a mimic might apply a new watermark to acquired works before fine-tuning mimicry. Our aim is for the fine-tuned model to still generate works with the original watermark, thereby protecting the artist's rights.

\section{Experiments}
\label{sec:exp}
\subsection{Experiment Setup}
\noindent\textbf{Dataset}.
We primarily evaluate the performance of the proposed method across three sets of image data:
\begin{itemize}
\item \textbf{Artworks from artists.} The study mainly used works from contemporary and historical artists (17 artists), each contributing about 50 paintings for key experiments. The contemporary artists have authorized their works for experiments and display.

\item \textbf{Artworks from text-to-models.}
We also evaluated our method's performance on images generated by 10 different AI models from the online art sharing website Civitai\cite{civitai}, with approximately 50 images for each model.

\item \textbf{Natural images.} We collected 10 sets of diverse natural images to assess our watermarking method, anticipating mimickers' use of real-world photographs for imitation.
\end{itemize}

\noindent\textbf{Model}.
The Stable Diffusion Model, used in our experiment for fine-tuning, is a leading generative model popular on websites and online communities.

\noindent\textbf{Watermark Methods}. We tested the performance of different watermarks in both general and multi-artist scenarios.
\begin{itemize}
\item \textbf{SSL \cite{fernandez2022sslwatermarking}.} SSL watermarking employs SSL-pretrained neural networks to embed watermarks in latent spaces. 

\item \textbf{RivaGAN \cite{zhang2019robust}.} RivaGAN is a robust image watermarking method using GANs with two adversarial networks for watermark assessment and removal.

\item \textbf{DWT-DCT \cite{al2007combined}.} 
The method enhances digital image watermarking by combining DWT with improved robustness and effectiveness.
\item \textbf{DWT-DCT-SVD \cite{he2018proposed}.} 
DWT-DCT-SVD method combines DWT, DCT, and SVD to watermark color images, converting RGB to YUV and embedding watermarks into DCT-divided, SVD-processed Y channel blocks.
\end{itemize}

\subsection{Metrics}
We assess the proposed method using bit accuracy rate (the percentage of correctly identified bits in the watermark), and in the context of mimicry fine-tuning, we also provide the best, average (avg), and overall accuracy distribution of the generated samples.

\subsection{Implement Details}
We use Stable Diffusion v1.5\cite{sdv1.5} as the base model. During the training phase, each image is standardized to a resolution of 512x512 pixels. The model undergoes 500 training steps for each image. The class tokens are uniquely combined with a rare identifier `zws' and a class prompt 'art\_paint'. In addition, we set the batch size to 2, use Lion as the optimizer, and the learning rate is 1e-07. In the generation phase, the model maintains the image size at 512x512 pixels. It utilizes 32 steps of sampling with the DDPM sampler, a scale setting of 7.5, and maintains the CLIP skip setting at 2. In our research, more than 100 experimental groups were trained using 4 RTX 3090 GPUs, collectively amounting to over 600 hours of GPU computation time. Experiments default to a 32-bit watermark length.

\begin{table*}[]
\centering
\begin{tabular}{lccccccc}
\hline
\multicolumn{1}{c}{\multirow{2}{*}{32-bit}} & \multicolumn{5}{c}{The sample numbers within an accurate bit range} & \multirow{2}{*}{avg(bits)} & \multirow{2}{*}{best(bits)} \\ \cline{2-6}
\multicolumn{1}{c}{}                       & 0-20\%    & 20-40\%    & 40-60\%   & 60-80\%   & 80-100\%   &                           &                            \\ \hline
no watermark                               & 0         & 86           & 908         & 6           & 0            & 15.18                     & 21                         \\ \hline
DWTDCT                                     & 0         & 43           & 502         & 439         & 16           & 17.41                     & 26                         \\
DWTDCTSVD                                  & 0         & 37           & 482         & 454         & 27           & 18.57                     & 30                         \\
SSL                                        & 0         & 0            & 237         & 748         & 15           & 19.29                     & 27                         \\
RivaGan                                    & 0         & 0            & 111         & 840         & 49           & 19.72                     & 29                         \\ \hline
\end{tabular}
\caption{Comparison of watermark extraction distribution across different watermarking methods in single-artist scenario (artist).\label{methods}}
\end{table*}

\subsection{Main Result}
\noindent\textbf{Single-Artist Fine-Tuning}.
Table~\ref{30} presents the efficiency of extracting 32-bit watermarks across various image types. Images without watermarks primarily exhibit extractions within the 40-60\% range, averaging around 14 bits. Conversely, watermarked images shift this average toward the higher 60-80\% range, with an approximate extraction of 19 bits. This trend persists among artist-created, AI-generated, and reality photo categories, suggesting that watermarking substantially increases the number of bits that can be extracted. Optimal extraction results vary by image category, achieving between 20 to 29 bits.

\begin{figure}[t]
\centering
\includegraphics[width=0.95\linewidth]{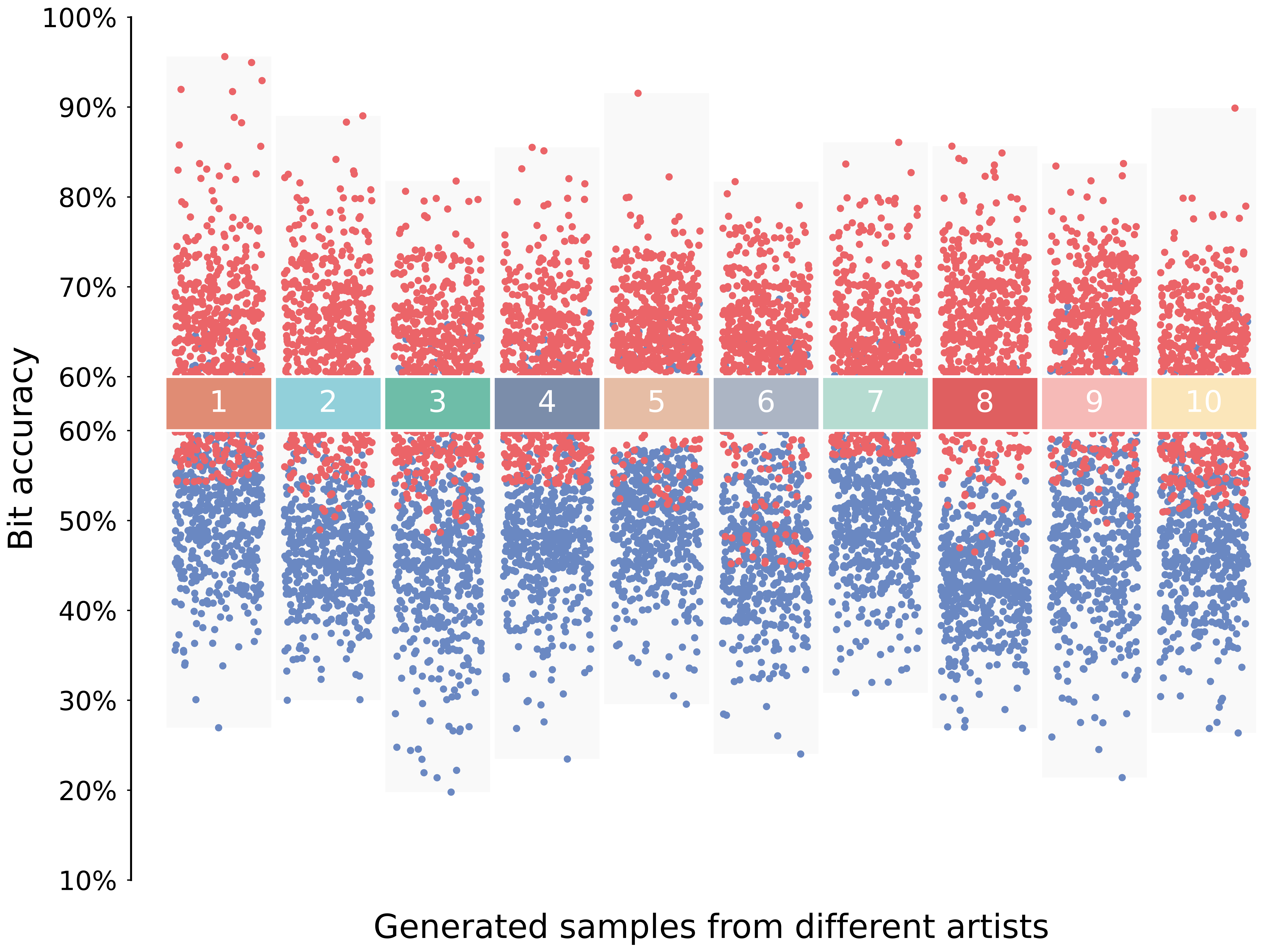}
       \caption{The watermark bit accuracy distribution of samples generated by the fine-tuning model with watermark (red) and without watermark (blue), 450 samples randomly selected for each set. }
   \label{10person}
\end{figure}

\begin{figure}[thb]
\centering
\includegraphics[width=0.95\linewidth]{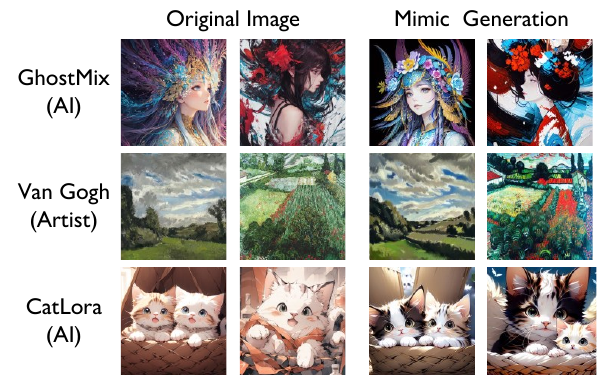}
       \caption{Original image (Left) and mimic generation from which watermark can be extracted (right). }
   \label{gs}
\end{figure}

Figure~\ref{10person} shows that fine-tuning models with stolen watermarked images leave watermark traces in the model, evidenced by a 60-80\% accuracy rate (red) in capturing these traces in generated outputs. This contrasts with the 40-60\% accuracy (blue) for models fine-tuned without watermarked images, demonstrating the latent watermark's detectability in generated styles. Figure~\ref{gs} displays the original image from GhostMix~\cite{ghost}, Van Gogh, and Catlora~\cite{cat} for fine-tuning, and the generated image with extractable watermark, maintaining visual quality through integrated mimic fine-tuning in the generation model.

\begin{figure*}[t]
  \centering
\subfloat[RivaGan]{\includegraphics[width=2.22in]{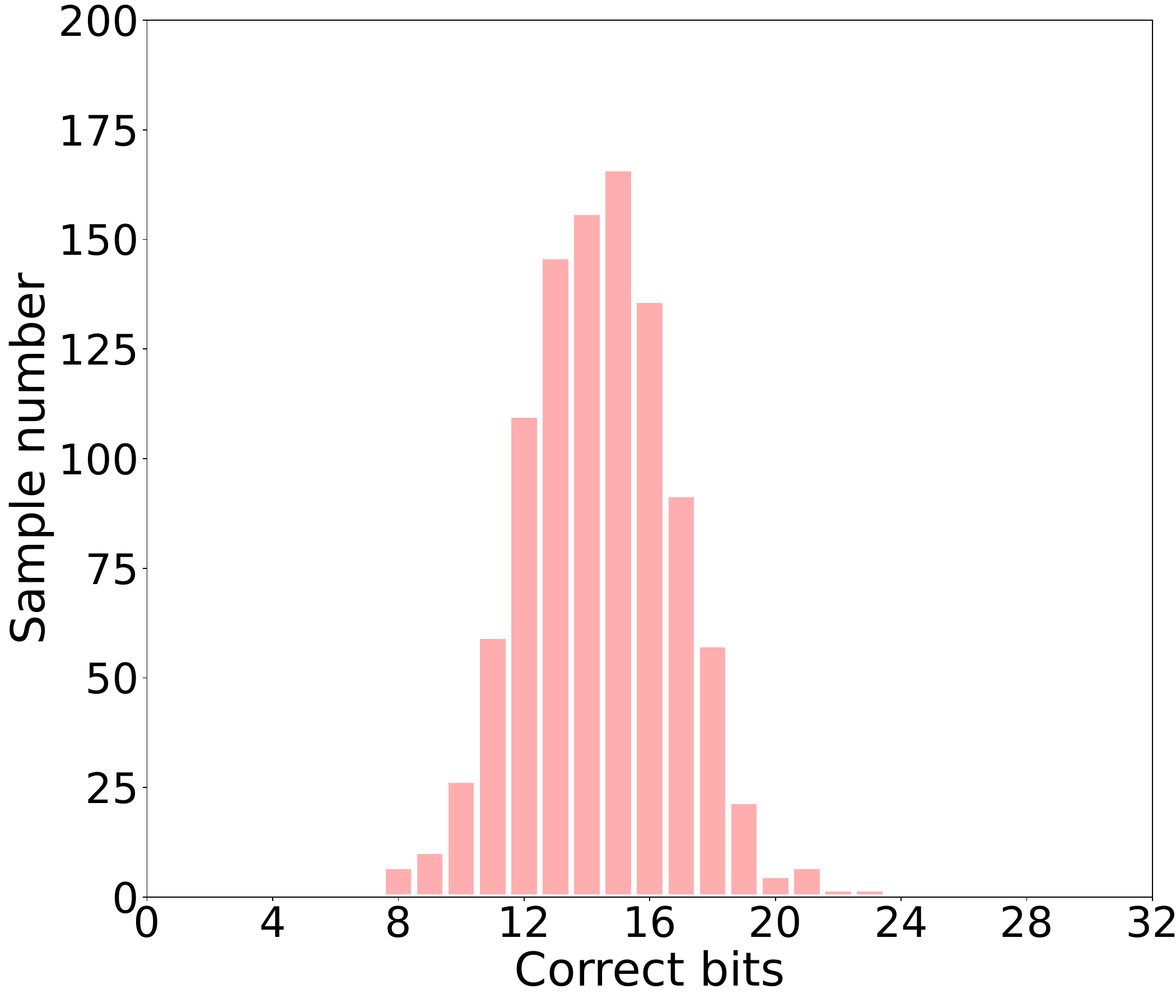}
\label{rivao}}
\hfil
\subfloat[SSL]{\includegraphics[width=2.22in]{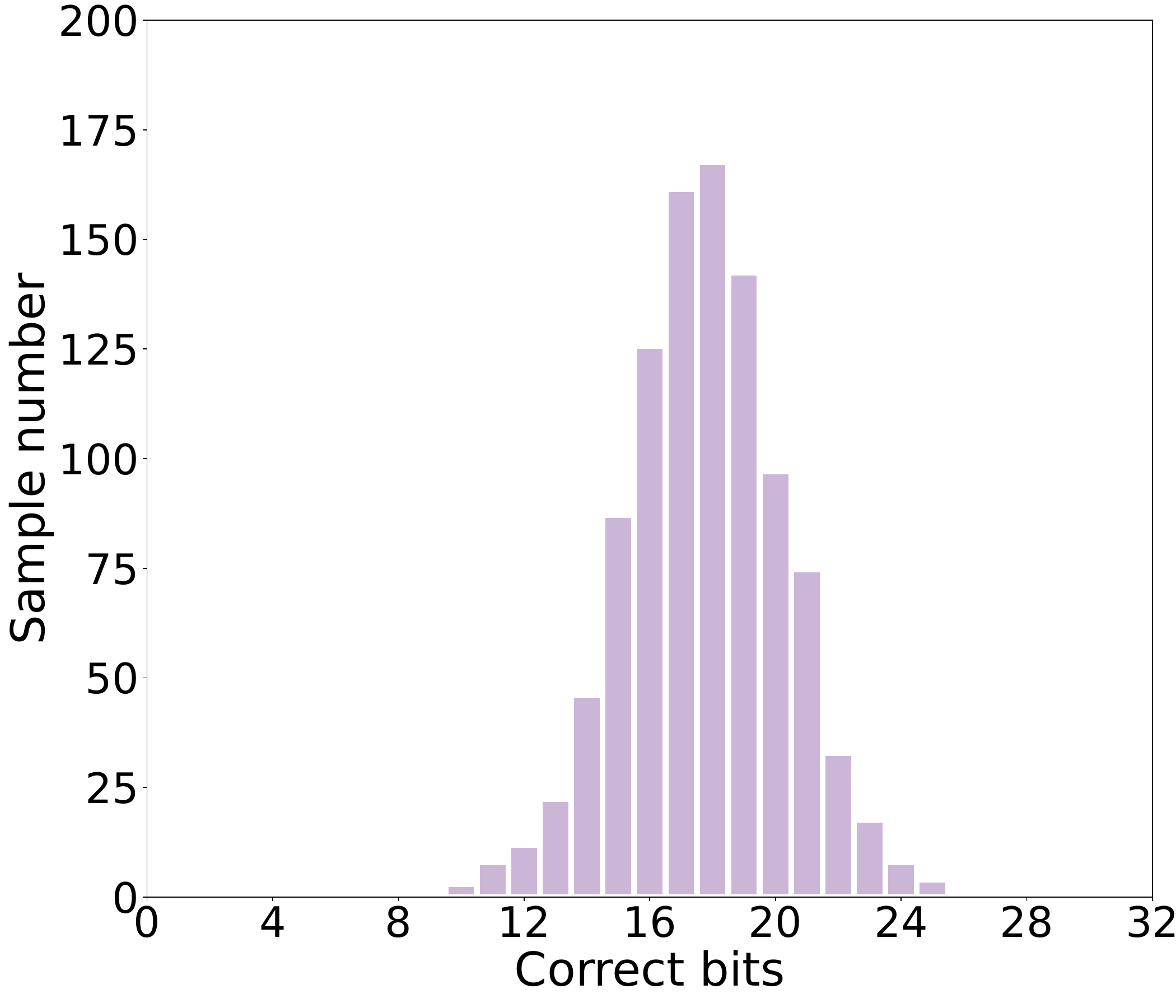}%
\label{sslo}}
\hfil
\subfloat[DWT-DCT-SVD]{\includegraphics[width=2.22in]{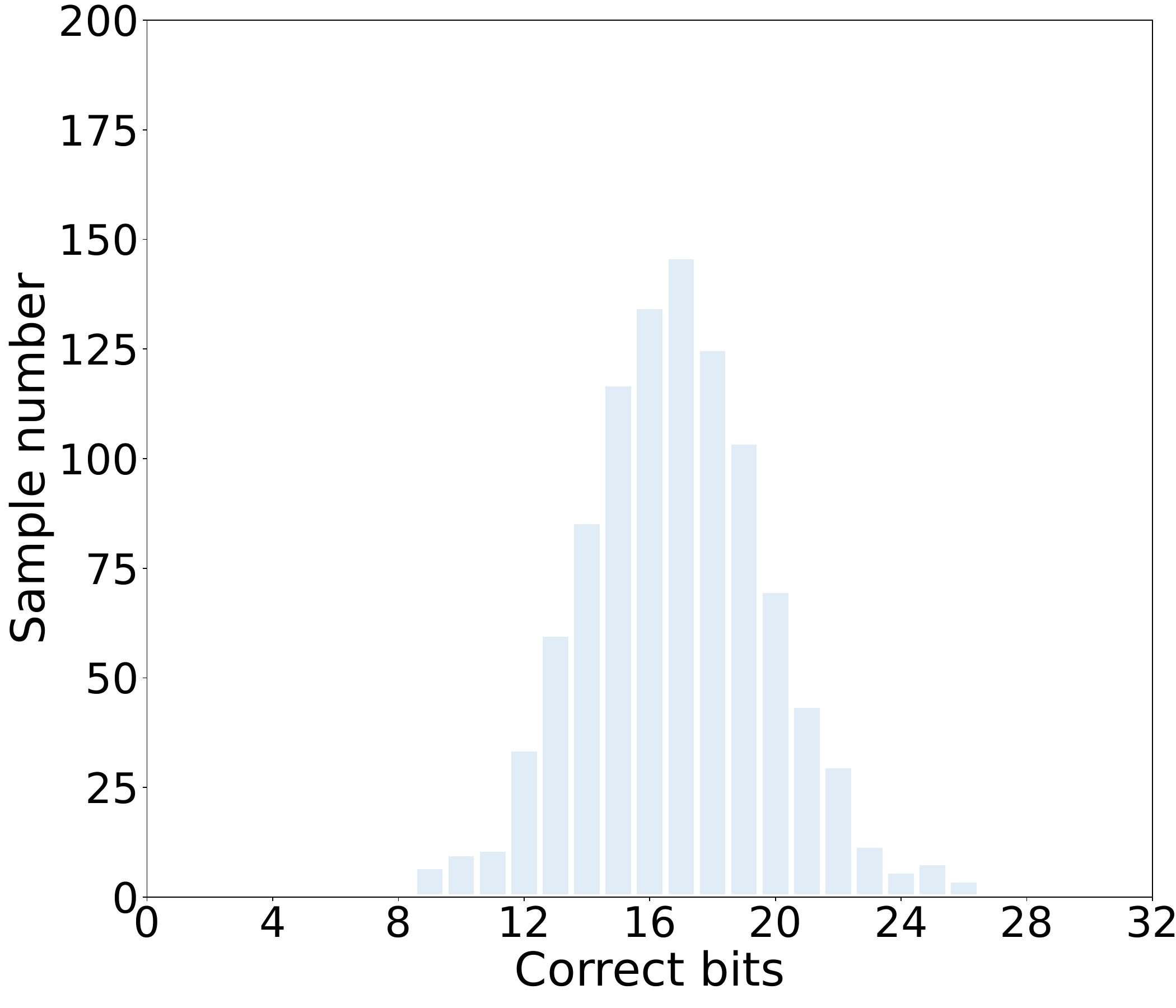}%
\label{ddso}}
\hfil
\caption{Fine-tuning of images with overlaid watermarks from three different methods and the extraction results for each watermark.\label{over}
}
\end{figure*}

\noindent\textbf{Different Watermark Methods}.
Table~\ref{methods} shows watermark extraction for different methods, with a clear trend: techniques like DWTDCTSVD and RivaGan achieve higher accuracy, mostly in the 60-80\% range, with top extractions up to 30 bits. This contrasts with the non-watermarked baseline, which peaks at 21 bits. Advanced methods demonstrate improved robustness in watermark retrieval.

\begin{figure*}[t]
  \centering
\subfloat[Same methods for 3-Artists mixed]{\includegraphics[width=2.22in]{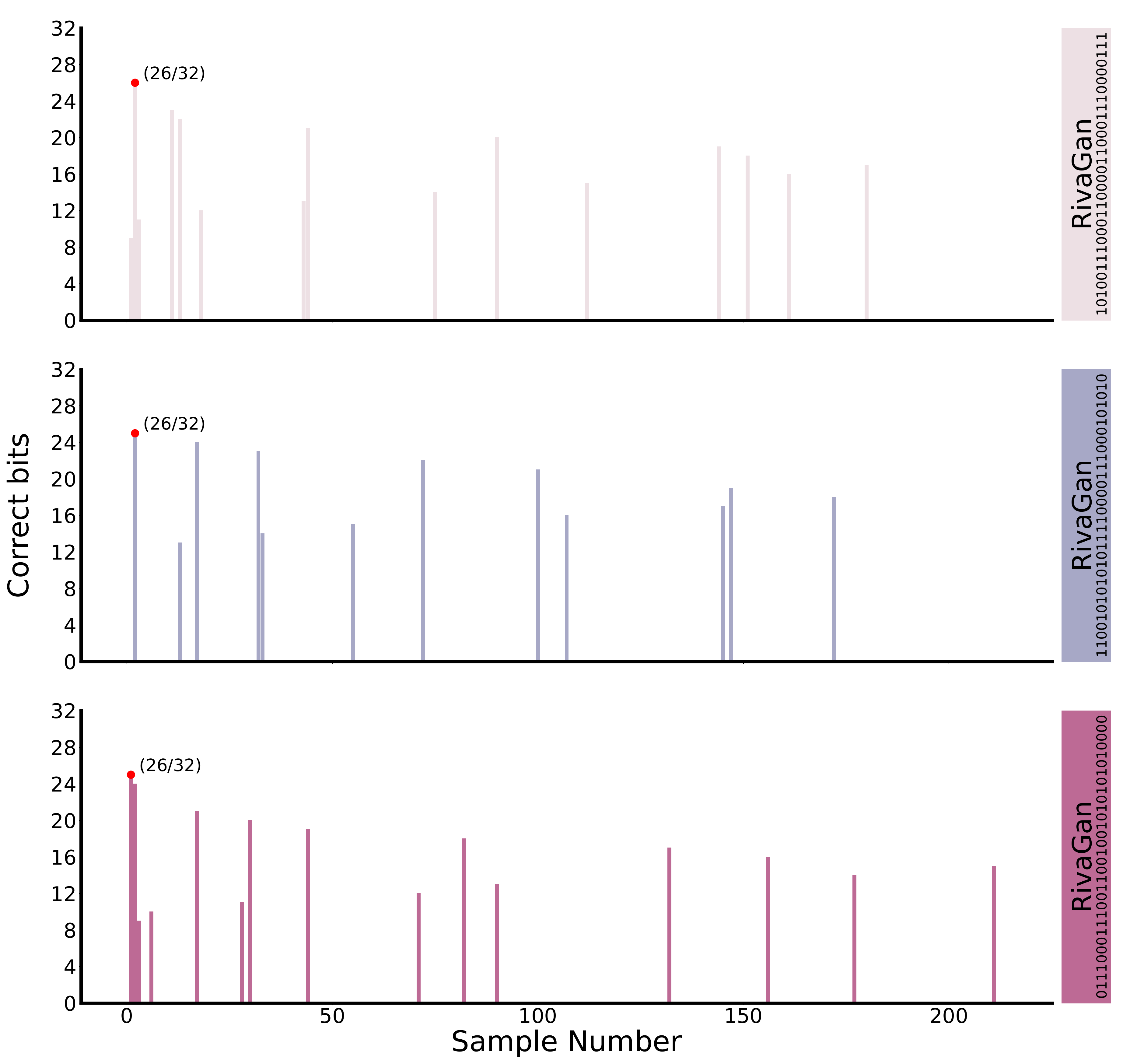}%
\label{same}}
\hfil
\subfloat[Differnt methods for 3-Artists mixed]{\includegraphics[width=2.22in]{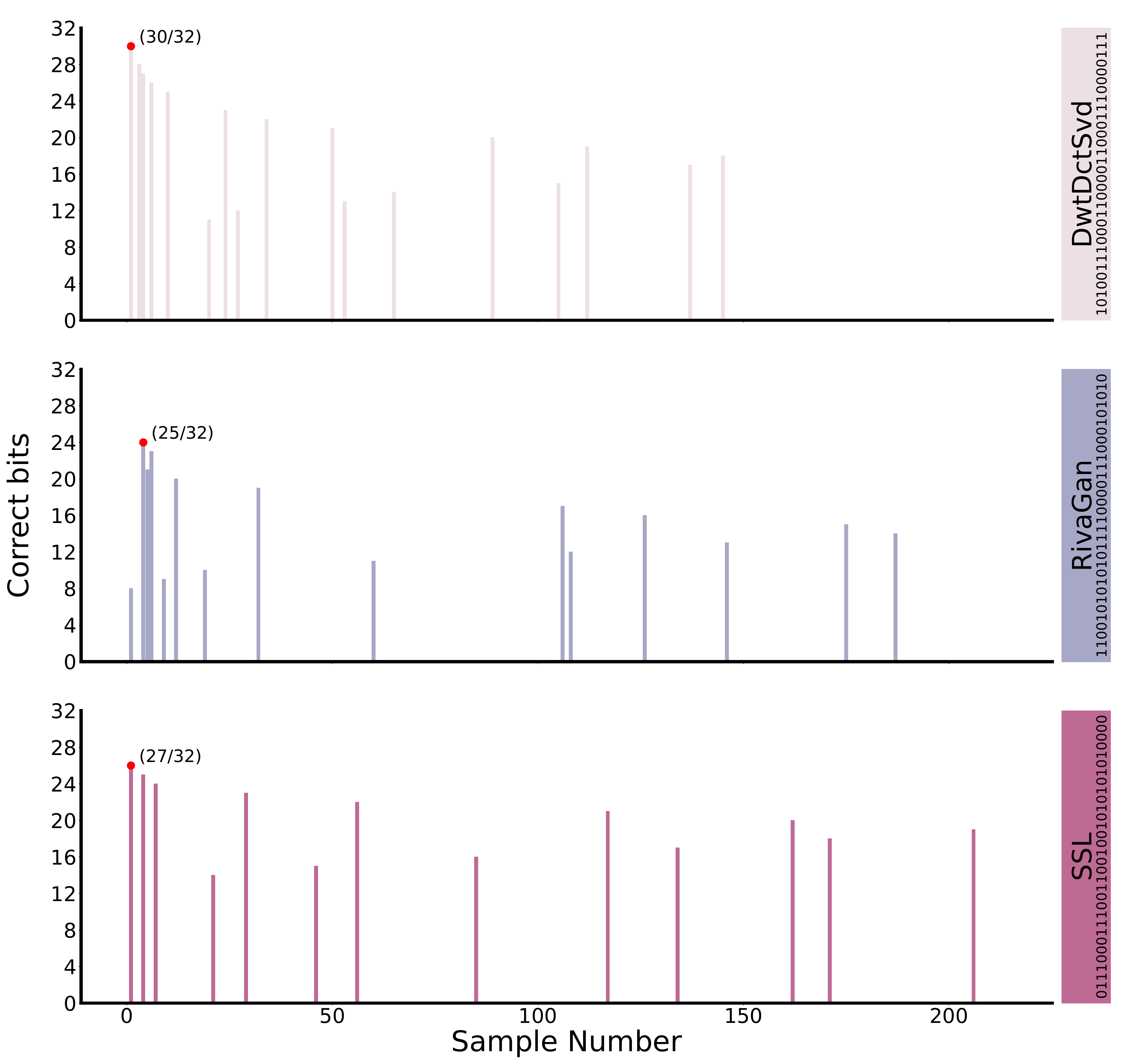}%
\label{differ}}
\hfil
\subfloat[Nowatermark for 3-Artists mixed]{\includegraphics[width=2.22in]{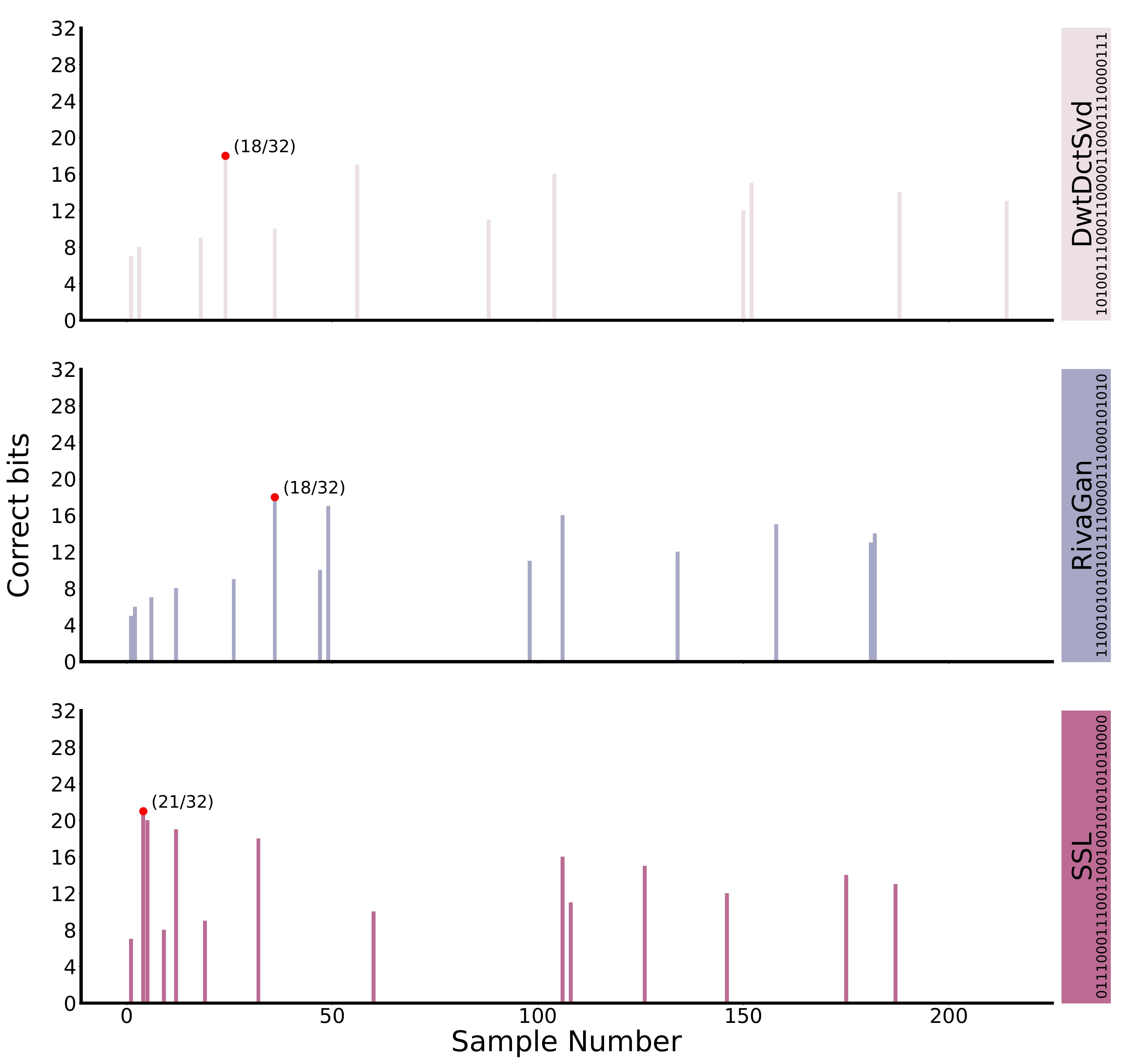}%
\label{nowa}}
\hfil
\caption{Multi-person (3 Artistss) mixed fine-tuning of watermark extraction scenarios: the same watermarking method, different methods, no watermarking.\label{mix}
}
\end{figure*}

\noindent\textbf{Multi-Artist Fine-Tuning}.
Figure~\ref{mix} illustrates the watermark extraction outcomes for artworks by three artists. They used a consistent watermarking method (RivaGAN), three distinct watermarking methods for individual works, and also a set without any watermarking. After mixed fine-tuning, these artworks were used to generate new pieces based on descriptions from the three artists. The results revealed that a single watermarking technique achieved a maximum accuracy of 26/32, whereas employing various methods resulted in a higher accuracy of 30/32. These results are significantly different from those where no watermarks were embedded.

\begin{figure*}[t]
  \centering
\subfloat[Variable-Proportion Watermarked Images.]{\includegraphics[width=2.22in]{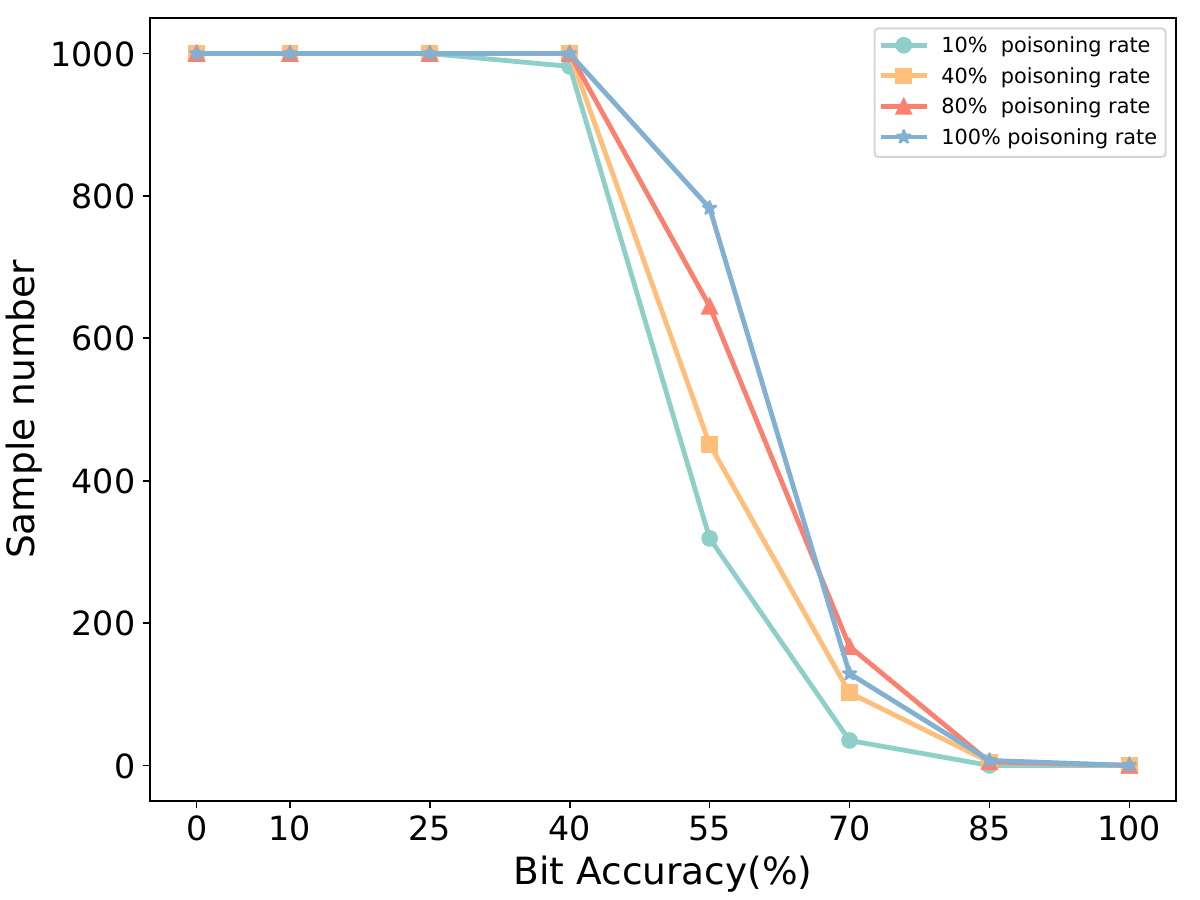}
\label{mr}}
\hfil
\subfloat[Variable-Bit Watermarks.]{\includegraphics[width=2.22in]{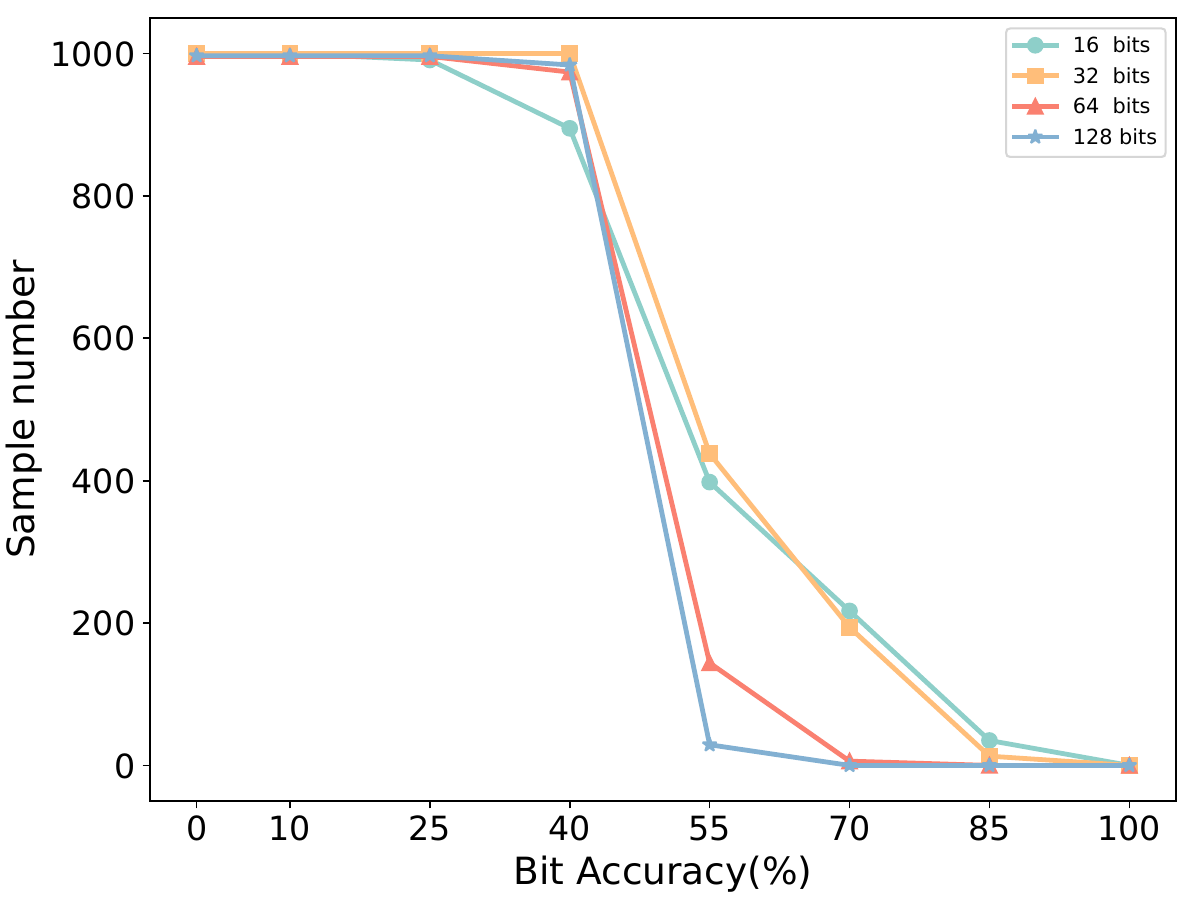}%
\label{mb}}
\hfil
\subfloat[Dreambooth and Lora for fine-tune]{\includegraphics[width=2.22in]{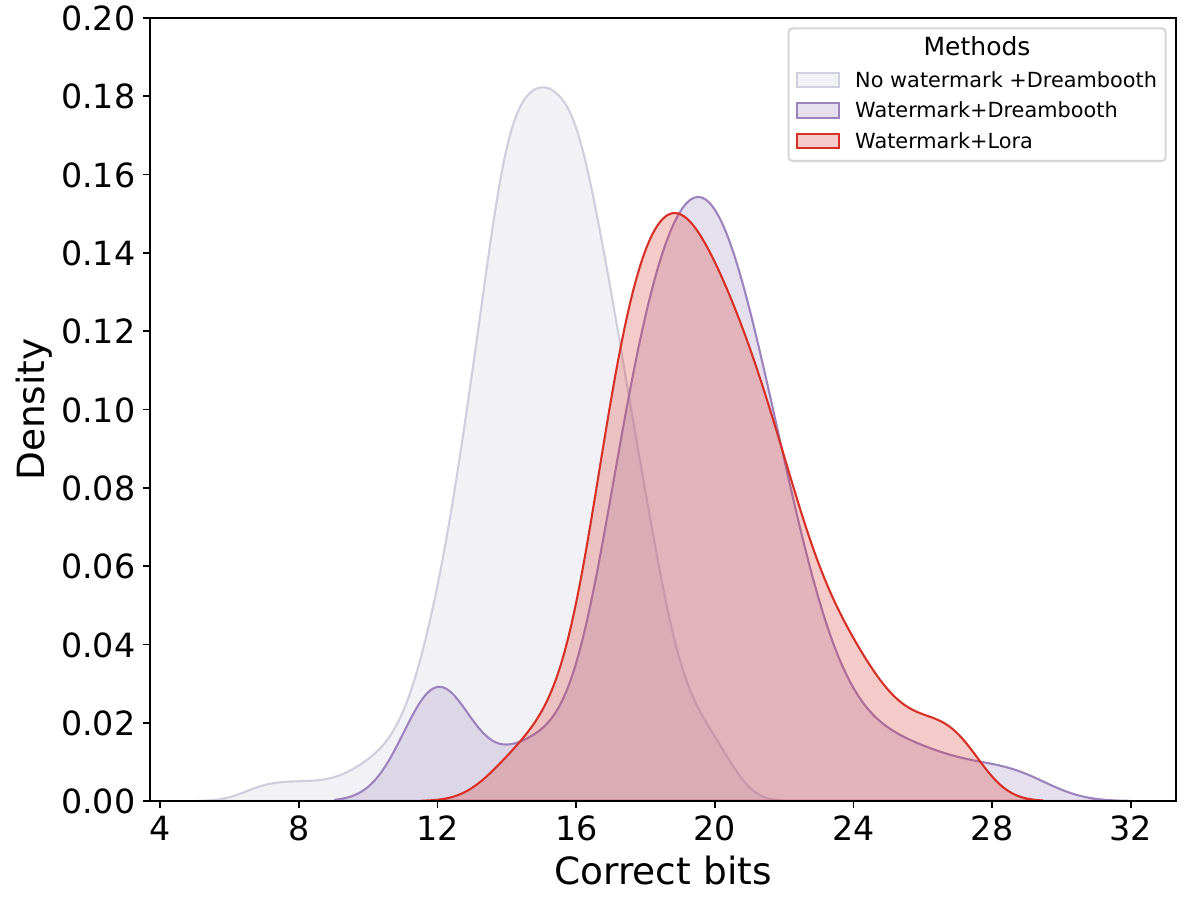}%
\label{lora}}
\hfil
\caption{Assessment of watermark extraction under mixed clean Fine-tuning conditions with varying proportions Watermarked images, evaluation of 16, 32, 64, 128-bit Watermark extraction, and Fine-tuning in differeant methods (Dreambooth and Lora).\label{other}
}
\end{figure*}

\noindent\textbf{Mixed Clean Fine-Tuning}.
Figure~\ref{mr} demonstrates watermark extraction efficiency from a varied set of clean images, fine-tuned with increasing proportions of watermarked images (10\%, 40\%, 80\%, and 100\% out of a total of 500). The figure highlights the variations in the number of samples with watermark bit extractions surpassing specific accuracy levels. It is observed that the extraction efficiency improves with a higher percentage of watermarked images. Remarkably, even with just 10\% of the images being watermarked, watermark extraction can achieve up to 85\% accuracy.

\begin{table*}[]
\resizebox{\linewidth}{!}{
\centering
\begin{tabular}{ccccccccccccc}
\hline
{  }                        & \multicolumn{10}{c}{{The sample numbers within an accurate bit range}}                                                                                                                                                                                                                                                               &                              &                           \\ \cline{2-11}
\multirow{-2}{*}{{32-bit}} &  { 0-10\%} & {10-20\%} & {20-30\%} & { 30-40\%} & {40-50\%} &    { 50-60\%}          & {60-70\%} & {70-80\%}                  & { 80-90\%}                  & {\ 90-100\%}                 & \multirow{-2}{*}{avg(bits)}        & \multirow{-2}{*}{best(bits)}    \\ \hline
Unwatermarked                                  & 0                             & 0                              & 2                              & 194                          & 532                            & 232                        & 40                             & 0                         & 0                        & 0                        & 15.25                        & 21                        \\ \hline
Normal Fine-tune                               & 0                             & 0                              & 0                              & 0                            & 0                              & 261                        & 640                            & 81                        & 13                       & 5                        & 19.96                        & 29                        \\ \hline
Two-stage Fine-tune                            & 0      &  0       & 0       &  11    & 125     & 245 & 571    &  44 & 4 & 0 &  19.02 & 26 \\
GaussianBlur                                   & 2                             & 0                              & 2                              & 30                           & 119                            & 211                        & 558                            & 67                        & 11                       & 0                        & 19.25                        & 28                        \\
Brightness                                     & 0                             & 0                              & 2                              & 33                           & 189                            & 294                        & 450                            & 30                        & 2                        & 0                        & 18.24                        & 26                        \\
CenterCrop                                     & 0                             & 0                              & 1                              & 49                           & 246                            & 279                        & 411                            & 14                        & 0                        & 0                        & 17.76                        & 24                        \\
Contrast                                       & 0                             & 0                              & 2                              & 55                           & 229                            & 266                        & 401                            & 44                        & 3                        & 0                        & 18.06                        & 26                        \\
Hue                                            & 0                             & 0                              & 0                              & 29                           & 226                            & 346                        & 372                            & 27                        & 0                        & 0                        & 17.92                        & 25                        \\
JPEG                                           & 0                             & 0                              & 0                              & 12                           & 171                            & 316                        & 473                            & 28                        & 0                        & 0                        & 18.45                        & 25                        \\
Meme                                           & 1                             & 0                              & 12                             & 115                          & 315                            & 263                        & 275                            & 18                        & 1                        & 0                        & 16.86                        & 26                        \\
Resize                                         & 0                             & 0                              & 0                              & 11                           & 148                            & 269                        & 488                            & 78                        & 6                        & 0                        & 19.08                        & 27                        \\
Rotation                                       & 0                             & 0                              & 0                              & 26                           & 165                            & 311                        & 460                            & 35                        & 3                        & 0                        & 18.85                        & 27                        \\ \hline
\end{tabular}}
\caption{Performance of watermark robustness against various attacks.
\label{4methods}}
\end{table*}

\begin{table}[]
\resizebox{\linewidth}{!}{
\begin{tabular}{ccc}
\hline
       Attempt to Revise the Prompt                                         & avg(bits) & best(bits) \\ \hline
Remove the word 'painting' from the caption     & 17.90     & 24         \\
Enrich the content of the caption               & 18.10     & 25         \\
Revise the grammatical structure of the caption & 19.12     & 26         \\
Replace with synonyms                           & 19.22     & 26         \\
Make minor modifications to the caption         & 19.32     & 26         \\
Remove adjectives from the caption              & 19.35     & 26         \\
Scramble the order of the caption               & 19.42     & 27         \\
Retain a single subject object                  & 20.28     & 27         \\
Simplify the caption                            & 20.29     & 27         \\
Add strange characters to the caption           & 21.06     & 27         \\ \hline
\end{tabular}}
\caption{Watermark extraction performance during style generation using different prompts.\label{prompt}
}
\end{table}

\subsection{Robustness Study}
\noindent\textbf{Under Traditional Attacks}.
Table \ref{4methods} and \ref{over} show that when images embedded with watermarks for SD fine-tuning are exposed to traditional attacks, like JPEG compression, the resultant generated images retain high bit accuracy. The average correct bits surpass those of the unmodified model, with a peak observed at 28 bits.

\noindent\textbf{Secondary Fine-Tuning}.
Table \ref{4methods} demonstrates that secondary SD fine-tuning on generated images retains a high bit accuracy. The result shows that it achieves a maximum accuracy of 26/32.

\subsection{Bit Length Study}
We assess the impact on watermark distribution in images from a stable diffusion model fine-tuned with watermarked images of various capacities. We use four watermark schemes: 16, 32, 64, and 128-bit. Figure \ref{mb} shows sample variations in bit accuracies in 1000 images generated by the fine-tuned SD model. Results indicate that samples with over 70\% accuracy for 64-bit and 128-bit schemes are almost negligible. However, with 16-bit and 32-bit schemes, over 200 samples achieved more than 70\% watermark accuracy, some reaching 85\%.

\subsection{Lora Study}
We evaluate the impact of various fine-tuning approaches on the stable diffusion model's ability to generate watermarked images. Figure \ref{lora} shows that the ‘No watermark(Dreambooth)’ distribution peaks at a lower bit accuracy (12-16 bits). In contrast, the distributions for 'Watermark (Dreambooth)' and 'Watermark (Lora)' are strikingly similar and exhibit a higher peak in bit accuracy, ranging from 19 to 24 bits.
\subsection{Prompt Study}
Given that the prompts used by the thieves to train with stolen artworks remain unknown, we modified 10 prompts to prompt the suspect model into generating images. As shown in Table \ref{prompt}, the average bit accuracy of 1000 images generated by the suspect model using each prompt type is significantly high, predominantly exceeding 19 bits. Notably, the peak average correct bits attained is 21.06 bits, with individual images reaching up to 27 bits. 
\section{Discussion and Conclusion}
\label{sec:discon}
\subsection{Discussion}
\noindent\textbf{Ethical Statement:} 
This paper features paintings with the direct authorization from the artists. Additionally, numerous contemporary artworks have been downloaded from social platforms solely for the purpose of style analysis and discussion, with no experimental usage prior to obtaining the artists' permission. To augment our data for experimental research, we collected historical artworks, AI style model-generated pieces, and various natural photos. We hereby affirm that this collected data is specifically for the experiments related to this paper's proposed methods and is not for any other use.

\noindent\textbf{Limitations:} 
The watermarking framework proposed in this study is conceptual and not sufficiently robust for partial scenarios. It struggles to extract watermarks precisely from a few samples for identity and authorization. When mimickers conceal their services and do not disclose their mimic models, sharing only a few style-generated images, the framework's ability to identify them is limited.

\subsection{Conclusion}
We introduce a watermarking scheme that integrates watermarks into artists' works to protect intellectual property while maintaining their art's aesthetic integrity. This system allows artists to track unauthorized art mimicry in fine-tuning applications. Our research covered various watermarking methods, fine-tuning conditions, and strategies against watermark compromise. The effectiveness of our approach is backed by substantial experimental evidence. Future efforts will focus on developing more robust watermarking techniques with higher extraction rates to counteract mimicry in fine-tuning.

{
    \small
    \bibliographystyle{ieeenat_fullname}
    \bibliography{main}
}


\end{document}